\documentclass{article}
\usepackage{ijcai25}

\usepackage{times}
\usepackage{soul}
\usepackage{url}
\usepackage[hidelinks]{hyperref}
\usepackage[utf8]{inputenc}
\usepackage[small]{caption}
\usepackage{graphicx}
\usepackage{amsmath}
\usepackage{amsthm}
\usepackage{booktabs}
\usepackage{algorithm}
\usepackage{algorithmic}
\usepackage[switch]{lineno}

\usepackage[normalem]{ulem}
\usepackage{multirow}
\usepackage[utf8]{inputenc}
\usepackage{CJKutf8}
\usepackage{hyperref}
\usepackage{xcolor}
\soulregister\cite7
\soulregister\ref7
\usepackage{ulem}
\usepackage{pgf-pie}
\usepackage{subcaption}

\title{Picturized and Recited with Dialects: A Multimodal Chinese Representation Framework for Sentiment Analysis of Classical Chinese Poetry}

\author{
Xiaocong Du\footnote{These authors contribute equally to this work.}
\and
Haoyu Pei\footnotemark[1]
\and
Haipeng Zhang\footnote{Corresponding author.} \\
\affiliations
ShanghaiTech University\\
\emails
\{duxc2023, peihy2024, zhanghp\}@shanghaitech.edu.cn
}

\begin{document}

\maketitle

\begin{abstract}
Classical Chinese poetry is a vital and enduring part of Chinese literature, conveying profound emotional resonance. Existing studies analyze sentiment based on textual meanings, overlooking the unique rhythmic and visual features inherent in poetry, especially since it is often recited and accompanied by Chinese paintings.
In this work, we propose a dialect-enhanced multimodal framework for classical Chinese poetry sentiment analysis. We extract sentence-level audio features from the poetry and incorporate audio from multiple dialects, which may retain regional ancient Chinese phonetic features, enriching the phonetic representation. Additionally, we generate sentence-level visual features, and the multimodal features are fused with textual features enhanced by LLM translation through multimodal contrastive representation learning.
Our framework outperforms state-of-the-art methods on two public datasets, achieving at least 2.51\% improvement in accuracy and 1.63\% in macro F1. We open-source the code to facilitate research in this area and provide insights for general multimodal Chinese representation.
\end{abstract}

\section{Introduction}
Classical Chinese poetry, as an essential component of Chinese literature, is marked by its concise expression, rich imagery, and structured tones and rhythms, conveying profound emotional resonance~\cite{liu2022art}. From ancient times to today, it remains widely spread and has profoundly influenced modern Chinese~\cite{zheng1995}. Sentiment analysis of classical Chinese poetry has gained increasing attention recently, with researchers building and refining models on benchmark datasets of manually labeled sentence-sentiment or poem-sentiment pairs~\cite{chen2019sentiment,wei2024knowledge}.

\begin{figure}[t]
	\centering
	\includegraphics[width=0.47\textwidth]{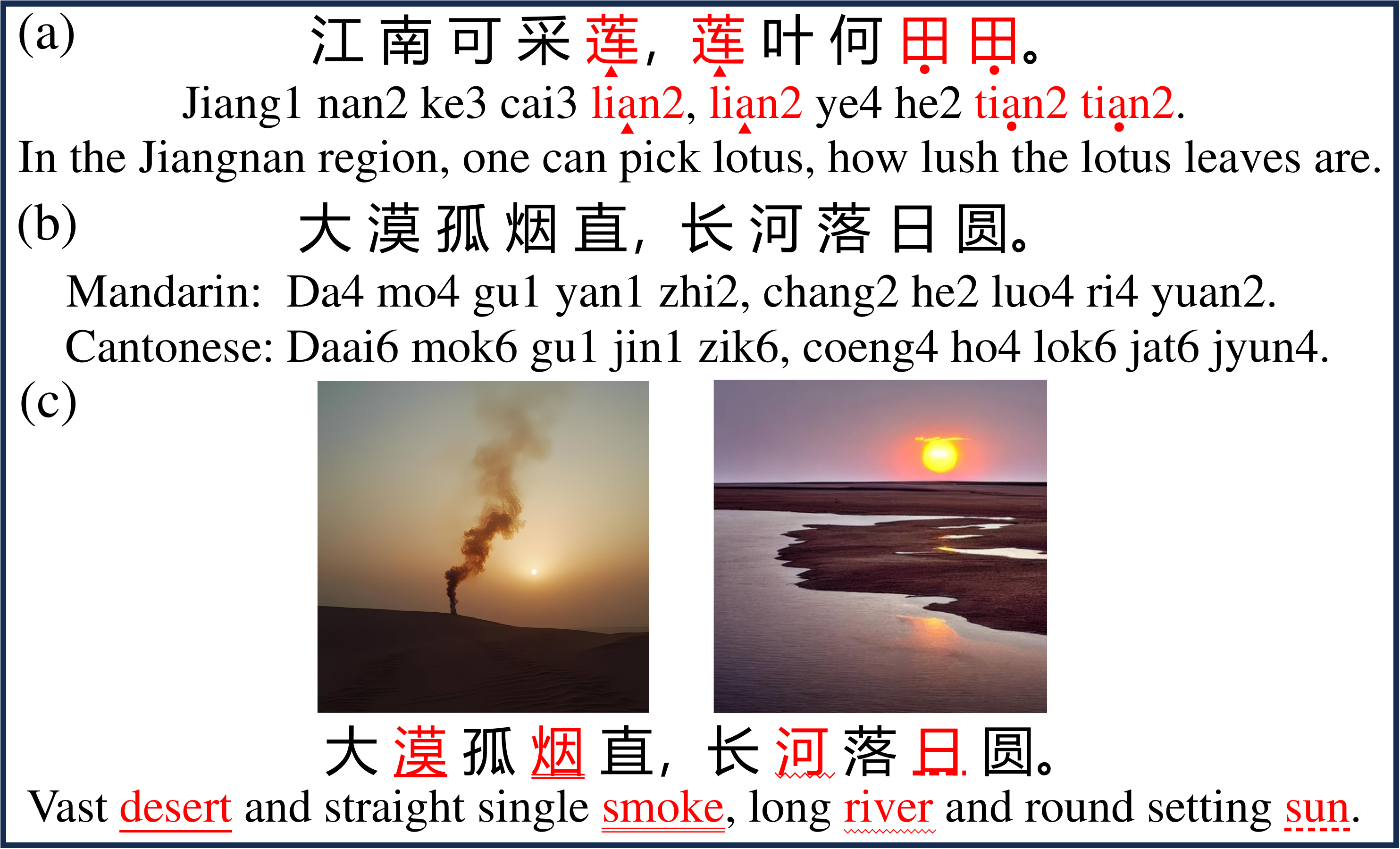}
	\caption{Case (a) illustrates the rhythmic features of classical Chinese poetry, Case (b) shows differences in dialectal pronunciations, and Case (c) demonstrates the visual features.} 
	\label{fig:intro}
\end{figure}

Existing research mainly focuses on textual analysis~\cite{wang2023rethinking,wei2024cross,xiang2024cross}, however, neglecting the fact that beyond the \textit{text} itself, poetry's emotions are in the meantime conveyed through the vivid \textit{picture} it evokes, and the \textit{rhythm} expressed in its recitation (example in Figure~\ref{fig:intro}). This is evidenced by the fact that classical Chinese poetry is often accompanied by Chinese paintings~\cite{qian1985}, and commonly recited~\cite{liu1985}, serving as a prevalent form of transmission.

We are developing a multimodal Chinese representation framework for sentiment analysis of classical Chinese poetry, while also hoping to deepen insights into general Chinese understanding. For this specific task, it addresses the problem of overlooking important visual and rhythmic information, as well as the lack of integration among these modalities. In the broader context of Chinese representation, most studies focus on multimodal fusion at the character level~\cite{meng2019glyce,sun2021chinesebert,mai2022pretraining}, while we represent semantics at the sentence level. This allows for the combination of different imagery and rhythmic elements to convey new and more complete information within the context of Chinese poetry. As a first effort, our framework also integrates dialects at the phonetic level, as their pronunciations may more closely resemble ancient forms from specific geographic areas, allowing for a more accurate representation of the emotional nuances. We elaborate on our motivations and design considerations as follows.

The \emph{rhythmic} features of poetry are reflected in the fact that poetry is specially designed for `recitation'~\cite{liu1985}. Strict phonological constraints, such as rhyming schemes and tonal patterns \begin{CJK}{UTF8}{gbsn}(平仄)\end{CJK}, govern the structure of classical Chinese poetry, allowing readers to experience emotions during poetry recitals. As shown in Figure~\ref{fig:intro}(a), \begin{CJK}{UTF8}{gbsn}in the poetic sentence `江南可采莲，莲叶何田田。', in addition to the usual end-line rhyming (`莲' lian2 and `田' tian2),\end{CJK} the same rhyming words appear four times in the sentence, creating a cyclical, repetitive tone when reciting that conveys a positive emotion.

The helpful rhythmic features may be influenced by variations in pronunciation across regions and over time, which can be reflected in \emph{dialects}. As shown in Figure~\ref{fig:intro}(b), some sentences may not adhere to the tonal patterns in Mandarin (based on northern dialects), while they may conform to those in Cantonese (a southern dialect), and vice versa. For a certain poem written in a certain time and space, there may be a more suitable dialect for reciting it, since some regional dialects can preserve ancient pronunciations~\cite{yuan2001}, necessitating the need to consider multiple dialects.

Meanwhile, ancient Chinese poets excelled at transforming objective nouns into picturized imagery, and the combination of imagery words constructs the \emph{visual} features of classical Chinese poetry~\cite{yuan2009}. \begin{CJK}{UTF8}{gbsn}For example, in Figure~\ref{fig:intro}(c), the sentence `大\uline{漠}孤\uuline{烟}直，长\uwave{河}落\dashuline{日}圆。' (Vast \uline{desert} and straight single \uuline{smoke}, long \uwave{river} and round setting \dashuline{sun}.) is composed solely of nouns and adjectives, while the picture created by these imagery words evokes in the reader a visual association, and the composition, hue, and other visual details of the picture lead to an emotional resonance.\end{CJK}

Considering these characteristics of this task, we design our model regarding rhythmic, dialect, and visual features. The rhythmic features of poetry are derived from the phonetic features of Chinese characters, which have been proven effective in general Chinese language understanding tasks~\cite{sun2021chinesebert,peng2021phonetic,mai2022pretraining}.
Previous work typically extracts phonetic features from the pinyin representation of Chinese characters, which is the letter-based phonetic representation. However, instead of pinyin, we use the audio of Chinese character pronunciations since it can provide fine-grained phonetic features~\cite{mai2022pretraining}. Besides, as poetry is meant to be recited for emotional delivery, it naturally motivates us to use audio rather than written symbols.
Furthermore, in previous methods, phonetic features are employed as auxiliary to enhance the token-level character representations. Since there are far more Chinese characters than pronunciations, the variances in characters capture much of the semantics, which may dominate the phonetic information, making it less expressive.
Given the critical role of phonetic features in our task, we propose to disentangle pronunciation embeddings from character embeddings and design a phonetic feature extraction module. The module ensures that phonetic information is more explicitly emphasized in the sentence-level features. Moreover, this disentangled design separates the audio modality from the text modality module, allowing our framework to be flexible and seamlessly integrate with other single-textual modality methods.

\begin{figure*}[ht]
	\centering
    \hspace*{0.2cm}
	\includegraphics[width=0.97\textwidth]{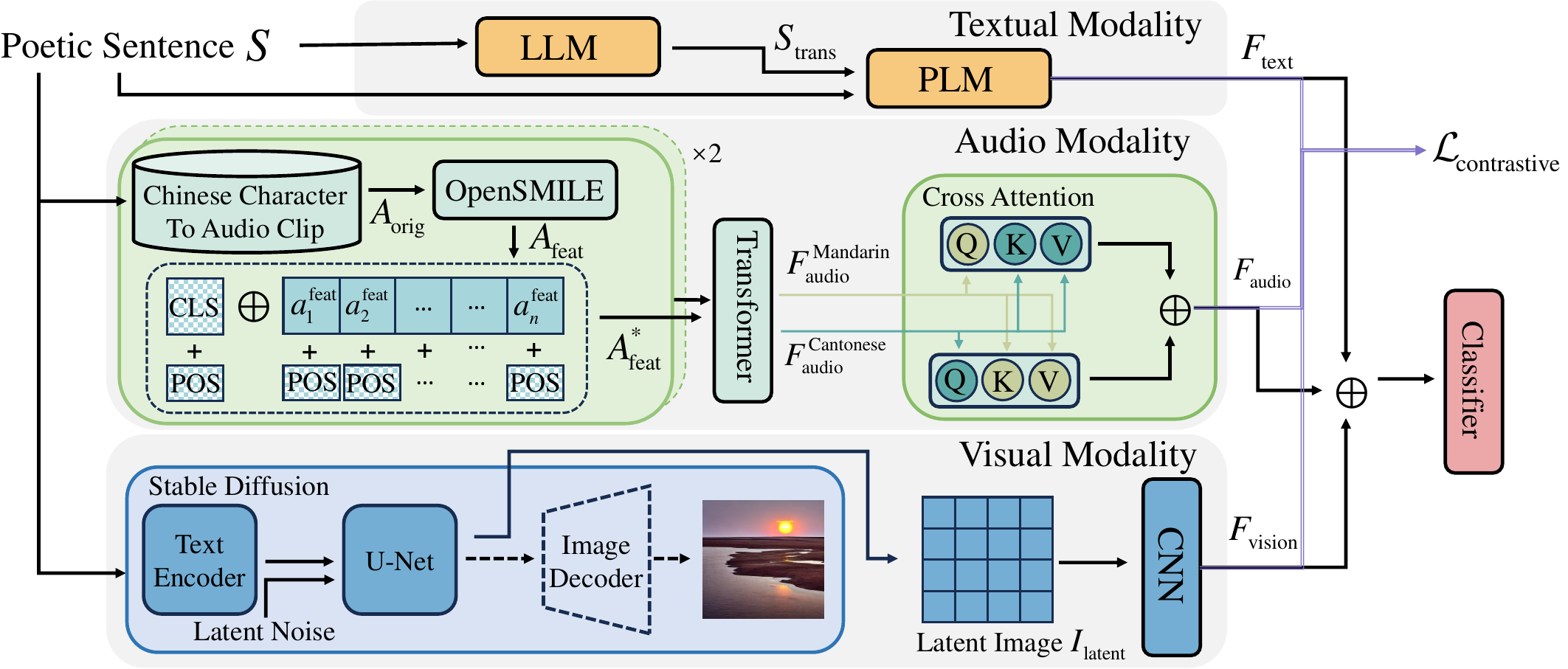}
	\caption{Tri-modal framework with textual (orange), audio (green), and visual (blue) modality modules. 
    PLM, Transformer, Cross-Attention, CNN, and Classifier are trainable. In the first pre-training phase (the purple line), we use contrastive loss $\mathcal{L}_{\text{contrastive}}$ to train the three modality feature extractors. In the second phase, all trainable modules are jointly trained on specific downstream tasks.}
	\label{fig1}
\end{figure*}

We then want to use dialect features. 
\begin{CJK*}{UTF8}{gbsn}Besides \textbf{Mandarin}, we choose four dialects as an additional audio source, including \textbf{Cantonese} (粤语), \textbf{Wu} (吴语), 
\textbf{Minnan} (闽南话), and \textbf{Chaozhou} (潮州话). We use Cantonese as an example below.\end{CJK*}We employ the same audio feature extraction module to obtain the Cantonese features and combine them with the Mandarin features. A challenge in audio feature fusion is that variations in speaker timbre and volume across different data sources inevitably introduce noise~\cite{chauhan2024enhancing}. Since noise is more likely to occur in the non-overlapping parts of the two features, our solution is to emphasize the shared features and pay less attention to others. We use cross-attention to align relevant features between two inputs by adjusting the attention weights~\cite{jian2024rethinking}, thereby fusing dialect audio features into the overall phonetic features.

The visual features presented in classical Chinese poetry are composed of imagery entities.
\citeauthor{su2023misc}~\shortcite{su2023misc} uses individual words from a poem as queries to find matching images from a search engine, extracting visual features from images to enhance the word embeddings. This approach is similar to looking up textual knowledge bases to enhance individual words~\cite{wei2024knowledge}. However, it does not provide a complete picture of the entire sentence and only offers information from isolated words.
Benefiting from recent advancements in text-to-image models~\cite{rombach2022high} trained on large-scale text-image pairs, our visual module can associate text with images to generate more complete visual representations at the sentence level.

Besides emphasizing phonetic and visual features, it remains essential to understand the concise and ancient textual content. We draw on prior work~\cite{xiang2024cross} and use a Large Language Model (LLM) to generate modern translations. Then we feed the original and translated texts into the pre-trained language model to obtain the overall text features before we integrate all three modality features. 

As shown in Figure~\ref{fig1}, we employ a two-stage training strategy to fuse audio (in green), visual (in blue), and textual (in orange) modality features. In the first stage, we train the three modality modules using multimodal contrastive representation learning~\cite{guzhov2022audioclip}. In the second stage, we concat the text, audio, and visual features to obtain the overall sentence-level features and train the entire framework.

We summarize our contributions as follows:
\begin{enumerate}
    \item We propose a tri-modal Chinese representation framework that fuses sentence-level audio, visual and textual features for classical Chinese poetry sentiment classification. Our method outperforms SOTA on two public datasets by 2.51\% in accuracy and 1.63\% in macro F1.
    \item This study is a first effort to explore the phonetic (audio) features of classical Chinese poetry for sentiment analysis. 
    Upon this, we incorporate features from four pairs of dialects into Chinese text representation through a cross-attention module. 
    We further confirm the effects of dialectal features with an empirical analysis.
    \item We demonstrate that our framework serves as a flexible tool, as popular text-only models for sentiment analysis achieve enhanced performance when integrated into it compared to their original capabilities. We open-source our framework\footnote{\url{https://github.com/ZhangDataLab/chinese_poetry_sentiment}} and hope it not only facilitates classical Chinese poetry sentiment analysis but also brings insights for general multimodal Chinese representation.
\end{enumerate}

\section{Related Work}

\subsection{Classical Chinese Poetry Sentiment Analysis}

A mainstream trend in sentiment analysis of classical Chinese poetry is to enhance the representation of poetic sentences with additional textual knowledge. CDBERT~\cite{wang2023rethinking} introduces dictionary definitions for rare words in poetry to improve the representation of words that appear less frequently during pretraining. SAMCAP~\cite{liu2024sentiment} extracts knowledge features from the knowledge base of imagery entities and sentiment words. Others use the syntactic structure of poetic sentences~\cite{wang2023enhancing,zhang2024confidence} and the radical components of Chinese characters~\cite{han2023rac}. 

Recent studies interpret classical Chinese using modern Chinese, which may not always be easily understood nowadays. Poka~\cite{wei2024knowledge} constructs a knowledge graph linking classical Chinese words with their modern Chinese equivalents. Other studies use large-scale classical-modern Chinese parallel corpora for pretraining, including different pretraining strategies, such as CCMC~\cite{liu2022contrastive}, CP-ChineseBERT~\cite{hong2023hybrid}, CGCLM~\cite{xiang2024cross} and CCDM~\cite{wei2024cross}.

However, most studies above focus on textual semantics, ignoring that classical Chinese poetry, which differs from other usual types of text, is also characterized by its rhythmic and visual features~\cite{liu2022art}. Although MISC~\cite{su2023misc} uses image features to enrich the imagery words, it still excludes the rhythmic features and does not fully integrate the phonetic, visual, and textual modalities. This motivates us to go beyond text to adopt a tri-modal perspective.

\subsection{Multimodal Chinese Representation}

Multimodal information can be used to enhance Chinese text representation. Many studies start with individual Chinese characters, which are the basic building blocks of the Chinese language. Glyce~\cite{meng2019glyce} is the first to employ a CNN-based model to extract visual features from the glyph images of Chinese characters. Building upon glyph embeddings, DISA~\cite{peng2021phonetic}  incorporates pronunciation embeddings. It employs a reinforcement learning framework to disambiguate the intonations of the Chinese characters. To improve the way that glyph and pronunciation features are fused, ChineseBERT~\cite{sun2021chinesebert} integrates character embeddings with glyph and pinyin embeddings into the BERT pretraining process and achieves SOTA performance on multiple Chinese NLP tasks.
Following this idea, MPM-CNER~\cite{mai2022pretraining} further introduces four pretraining tasks specifically designed for this task.

However, unlike previous studies focusing on representation at the character level, we extract the representation of the sentence semantics. The rhythmic and visual characteristics of classical Chinese poetry make it inherently suitable for extracting sentence-level features. This idea may also be extended to texts with similar characteristics to classical Chinese poetry.

\subsection{Multimodal Sentiment Analysis}

The broad sentiment analysis has expanded from text classification to include other modalities such as images and audio. This evolution is aligned with the increasing availability of multimodal data, such as image-text comments on social media~\cite{zhu2022multimodal} and audiovisual content on video platforms~\cite{hazarika2020misa}. In most multimodal sentiment analysis studies, the data from different modalities naturally coexist, leading to a focus on aligning and fusing features across modalities~\cite{gandhi2023multimodal}. In our work, however, we emphasize that we ``create'' multiple modalities from the original single text modality data.
We extend the input from a single textual modality to a multimodal approach, allowing for a multi-perspective understanding. This involves extracting audio features from the Chinese character audio and obtaining visual features using a text-to-image model.

\section{Method}

The overall structure of our model is shown in Figure~\ref{fig1}. First, we generate tri-modal features for the given poetic sentences. For the \textbf{audio modality} (3.1), we design a sentence-level audio feature extraction module to capture dialect-specific phonetic features. These dialectal features are then fused into a unified phonetic feature using a cross-attention module (3.2). For the \textbf{visual modality} (3.3), we extract intermediate features from a text-to-image model and process them with an image feature extractor to obtain the visual representation of the poetic sentences. For the \textbf{textual modality} (3.4), we employ LLM to generate modern Chinese translations of classical Chinese poetry. These translations are concatenated with the original text and jointly trained using a pre-trained language model. Finally, the features from all three modalities are fused through a warm-up training phase guided by multimodal contrastive representation learning (3.5), and the fused features are used for sentiment classification.

\subsection{Audio Modality Feature Extraction}

Commonly used audio feature extraction models, such as wav2vec2.0~\cite{baevski2020wav2vec} and AST~\cite{gong21b_interspeech}, are trained on complete, real audio recordings. While we do not have such recordings for the benchmark datasets, the audio of the sentences is alternatively composed of multiple character audio segments, making conventional audio feature extractors unsuitable. We therefore design a feature extraction model tailored for segment-level audio sequences. We first collect Mandarin audio clips\footnote{\url{https://chinese.yabla.com/chinese-pinyin-chart.php}} corresponding to individual Chinese characters. For a given poetic sentence $S=[c_1, c_2, \ldots, c_n]$, we map each character $c_i$ to its corresponding audio clip $a_i^{\text{orig}}$, resulting in an audio sequence $A_{\text{orig}}=[a_1^{\text{orig}}, a_2^{\text{orig}}, \ldots, a_n^{\text{orig}}]$. Then, we use OpenSMILE~\cite{eyben2010opensmile} to extract acoustic features from each clip, obtaining a token-level audio feature sequence $A_{\text{feat}}=[a_1^{\text{feat}}, a_2^{\text{feat}}, \ldots, a_n^{\text{feat}}]$. To capture the rhyming position and tonal patterns of specific characters in the poetic sentence, we add position embedding $p_i$ to the audio features of each token $a_i^{\text{feat}}$. Additionally, a class embedding $a^{\text{cls}}$ is appended at the beginning of the audio feature sequence $A_{\text{feat}}$ to aggregate the features of the entire sequence. Both position embedding and class embedding are learnable parameters. The final audio feature sequence $A_{\text{feat}}^* = [a^{\text{cls}} + p_0, a_1^{\text{feat}} + p_1, \ldots, a_n^{\text{feat}} + p_n]$ is sent to a transformer encoder to extract sentence-level phonetic features $F_{\text{audio}}^{\text{Mandarin}}$:

\begin{equation}
	F_{\text{audio}}^{\text{Mandarin}}=\text{Transformer} \left(A_{\text{feat}}^*\right)
\end{equation}

\subsection{Dialect Feature Fusion}
Mandarin (Putonghua) is based on northern dialects, while we also include Cantonese, a major southern dialect that may contain distinct regional and ancient phonetic features.
In addition to Mandarin phonetic features $F_{\text{audio}}^{\text{Mandarin}}$, we collect Cantonese audio clips\footnote{\url{https://humanum.arts.cuhk.edu.hk/Lexis/lexi-can/}} and obtain their features $F_{\text{audio}}^{\text{Cantonese}}$ in the same way. The audio feature extractors for both dialects share the same parameters, aiming to map features from different dialects into a shared vector space. We then want to fuse their phonetic features. However, differences in voice characteristics among speakers in different audio sources can introduce noise. To mitigate this, we introduce a cross-attention module that enhances overlapping features across different audio sources while minimizing divergent features. We first transform the Mandarin phonetic features $F_{\text{audio}}^{\text{Mandarin}}$ into the query $Q^{\text{Mandarin}}$, key $K^{\text{Mandarin}}$ and value $V^{\text{Mandarin}}$:

\begin{align}
	Q_{\text{Mandarin}} &= \text{Linear}_Q\left(F_{\text{audio}}^{\text{Mandarin}}\right) \\
    K_{\text{Mandarin}} &= \text{Linear}_K\left(F_{\text{audio}}^{\text{Mandarin}}\right) \\
    V_{\text{Mandarin}} &= \text{Linear}_V\left(F_{\text{audio}}^{\text{Mandarin}}\right)
\end{align}

For the Cantonese phonetic features $F_{\text{audio}}^{\text{Cantonese}}$, we use the same method to obtain the query $Q_{\text{Cantonese}}$, key $K_{\text{Cantonese}}$ and value $V_{\text{Cantonese}}$. Subsequently, the query and key from different dialects are used to calculate dot-product attention to focus on the common features within the audio. This process is expressed as:

\begin{align}
	F_{\text{audio-att}}^{\text{Mandarin}} &= \text{Softmax}\left(\frac{Q_{\text{Cantonese}}K_{\text{Mandarin}}^T}{d}\right)V_{\text{Mandarin}} \\
    F_{\text{audio-att}}^{\text{Cantonese}} &= \text{Softmax}\left(\frac{Q_{\text{Mandarin}}K_{\text{Cantonese}}^T}{d}\right)V_{\text{Cantonese}}
\end{align}

Finally, the enhanced features from different dialects are concatenated to obtain the audio modality features $F_{\text{audio}}$.

\begin{equation}
	F_{\text{audio}}=\text{Linear}_{\text{audio}} \left(\left[F_{\text{audio-att}}^{\text{Mandarin}} \oplus F_{\text{audio-att}}^{\text{Cantonese}}\right]\right)
\end{equation}

\subsection{Visual Modality Feature Extraction}

The visual modality features are extracted from a text-to-image model. We use the Chinese stable diffusion model `Taiyi'~\cite{zhang2022fengshenbang} with frozen parameters. The model consists of three components: a text encoder, an image information creator (U-Net), and an image decoder. The poetic sentence $S$ is first input to the text encoder. Subsequently, the token embeddings and a latent noise tensor are fed into the U-Net, resulting in a latent image information tensor $I_{\text{latent}}$:

\begin{equation}
	I_{\text{latent}}=\text{U-Net} \left(\text{TextEncoder}\left( S \right), \text{Latent Noise} \right)
\end{equation}

We choose not to use the image decoder, as we do not require the generated images but only the intermediate visual features of the latent space. The intermediate features directly represent the high-level correspondence between text and images. These features preserve the core imagery described in the poetry while reducing the information loss or pixel-level noise introduced by image decoding.

We then use a three-layer convolutional network to convert the latent image $I_{\text{latent}}$ into visual modality features $F_{\text{vision}}$:

\begin{equation}
	F_{\text{vision}}=\text{CNN} \left(I_{\text{latent}}\right)
\end{equation}

\subsection{Textual Modality Feature Extraction}

Understanding the text of the poetic sentence is also essential. Motivated by recent work in utilizing the alignment between classical Chinese and modern Chinese~\cite{xiang2024cross} for better comprehension of classical Chinese texts, we prompt\footnote{The prompt: `Translate the following poem into modern Chinese and provide a detailed explanation of what it describes.'} LLM GLM-4\footnote{\url{https://bigmodel.cn/dev/api/normal-model/glm-4}} to generate modern Chinese translations and explanations for classical Chinese poetry:

\begin{equation}
	S_{\text{trans}}=\text{LLM} \left(S,\text{prompt}\right)
\end{equation}

Subsequently, the original text and its translation are fed into GuwenBERT\footnote{\url{https://github.com/ethan-yt/guwenbert}}, a classical Chinese pre-trained language model, obtaining the final textual modality features $F_{\text{text}}$:

\begin{equation}
	F_{\text{text}}=\text{PLM} \left(S,S_{\text{trans}}\right)
\end{equation}

\subsection{Multimodal Feature Fusion}

Since our textual modality model is pre-trained while the audio and visual modality feature extractors are trained from scratch, the performance imbalance between them can lead to suboptimal learning efficiency in the overall framework's training. To address this, we implement a two-stage training strategy. In the first stage, we perform a warm-up training for the feature extractors of all three modalities using multimodal contrastive representation learning~\cite{guzhov2022audioclip}, which aligns the modalities through a self-supervised approach. The training loss for the first stage is:

\begin{align}
	\mathcal{L}_{\text{contrastive}} = & \mathcal{L}_{\text{cs-ce}} \left(F_{\text{audio}},F_{\text{vision}}\right) + \mathcal{L}_{\text{cs-ce}} \left(F_{\text{audio}},F_{\text{text}}\right) \nonumber \\ 
    & + \mathcal{L}_{\text{cs-ce}} \left(F_{\text{vision}},F_{\text{text}}\right)
\end{align}

Here $\mathcal{L}_{\text{cs-ce}}$ is the cosine similarity contrastive loss~\cite{radford2021learning}. It refers to calculating the cosine similarity between every pair of samples within a batch and then performing a binary classification for each pair to determine whether they match.

In the second stage, the features from three modalities are concatenated to form an overall feature representation of the poetic sentence, which is then classified using a linear layer. 

\begin{equation}
	\hat{y}=\text{Softmax} \left(\text{Linear}_{\text{fusion}}\left(\left[F_{\text{text}} \oplus F_{\text{audio}} \oplus F_{\text{vision}}\right]\right)\right)
\end{equation}

For single-label classification tasks, we use cross-entropy loss $\mathcal{L}_{\text{CE}}$, and for multi-label classification tasks, we use binary cross-entropy loss $\mathcal{L}_{\text{BCE}}$. The training loss for the second stage on different downstream tasks is:

\begin{align}
	\mathcal{L}_{\text{CE}} &= -\frac{1}{N} \sum_{i=1}^{N} y_i \log(\hat{y}_i) \\
    \mathcal{L}_{\text{BCE}} &= -\frac{1}{N} \sum_{i=1}^{N} \sum_{j=1}^{M} \left[ y_{ij} \log(\hat{y}_{ij}) + (1 - y_{ij}) \log(1 - \hat{y}_{ij}) \right]
\end{align}

\begin{table*}[h]
	\centering
    \small
	\begin{tabular}{cccccccc}
		\toprule
		Dataset & Text & Task & Samples & Categories & Train & Val & Test \\
		\midrule
		FSPC & Sentence & Single-label classification & 20,000 & 5 & 16,000 & 2,000 & 2,000 \\
		CCPD & Whole-Poem & Multi-label classification & 17,103 & 12 & 11,971 & 3,420 & 1,712 \\
		\bottomrule
	\end{tabular}
	\caption{Details of the FSPC and CCPD datasets.}
	\label{table:data_statistics}
\end{table*}

\section{Experiments}

\subsection{Experimental Setup}

\paragraph{Datasets} We evaluate our model using two public datasets. The THU Fine-grained Sentimental Poetry Corpus (\textbf{FSPC}) dataset~\cite{chen2019sentiment} is designed for a single-label classification task with five categories. It consists of 5,000 poems, each containing four sentences. Every sentence is assigned a sentiment label from one of five categories: `negative', `implicit negative', `neutral', `implicit positive', and `positive'. 

The second is the Chinese Classical Poetry Dataset (\textbf{CCPD})~\cite{wei2024knowledge} for multi-label classification with 12 categories, which contains 17,103 poems. Each poem has multiple whole-poem-level labels. The task has 12 categories, including `homesick', `frustrated', etc. See details in Table~\ref{table:data_statistics}.

\paragraph{Train/Test Split} The split is consistent with previous studies to ensure a fair comparison. As detailed in Table~\ref{table:data_statistics}, FSPC is divided into training, validation, and test in a ratio of 8:1:1, while for CCPD, the ratio is 7:2:1.

\paragraph{Implementation Details} We use the base-sized GuwenBERT to obtain text embeddings. The audio feature extractor is a 3-layer transformer with hidden size $d=768$ and attention heads $h=8$. The visual feature extractor is a 3-layer CNN. We train the model using the Adam optimizer with a learning rate of $1e^{-6}$. We conduct our experiments with two RTX 3090.

\paragraph{Evaluation Metrics} Being consistent with previous work, we use accuracy (Acc) as the evaluation metric on FSPC. For CCPD, we use micro F1 (Mic-F1) and macro F1 (Mac-F1).

\paragraph{Baselines} Due to the availability of the code for the baselines in the original papers that work on the two datasets, we present the results from those papers. Each dataset includes six baselines, with only one baseline shared between them, resulting in a total of 11 baselines.

We begin by comparing our method with ChatGLM, a commonly used Chinese Large Language Model, on both the FSPC and CCPD datasets. On the FSPC dataset, we further compare it with two knowledge-enhanced models, CDBERT and RAC-BERT. The other three baselines, CCMC, CCDM, and CGCLM, use pretraining on large-scale classical-modern Chinese parallel corpora. 

On the CCPD dataset, besides the fine-tuned ChatGLM, we evaluate two traditional multi-label classification models, HiAGM-TP~\cite{zhou2020hierarchy} and LCM~\cite{guo2021label}, as well as two knowledge-enhanced text classification models: GreaseLM~\cite{zhang2022greaselm}, KPT~\cite{hu2022knowledgeable}, and a modern-Chinese-enhanced method Poka. 

Meanwhile, we test whether our framework can enhance the single textual modality models for this task. We do this by applying an open-source textual model first and comparing its result with our framework that uses this specific textual model as its textual modality. The models include RoBERTa~\cite{cui-etal-2020-revisiting}, ERNIE~\cite{sun2021ernie}, ChineseBERT~\cite{sun2021chinesebert}, and SikuRoBERTa~\cite{dongbo2021sikubert}.

All experiments are conducted using base-sized models. We briefly review some of the key baselines that previously achieved SOTA performance on classical Chinese poetry sentiment classification:

\begin{itemize}
    \item \textbf{ChatGLM~\cite{du2022glm}:} ChatGLM is a Large Language Model fine-tuned for Chinese, demonstrating outstanding performance across various natural language processing tasks.
    \item \textbf{CDBERT~\cite{wang2023rethinking}:} CDBERT incorporates external dictionary definitions for rare words and polysemous words, addressing the limitations of conventional pre-trained language models in modeling such words.
    \item \textbf{RAC-BERT~\cite{han2023rac}:} RAC-BERT introduces two radical-aware pre-training tasks to enhance the token-level representation of Chinese characters.
    \item \textbf{CCMC~\cite{liu2022contrastive}:} The Chinese Classical and Modern Corpus Model (CCMC) performs pre-training by replacing classical Chinese words with their modern counterparts and employs contrastive learning to bridge the language gap.
    \item \textbf{CCDM~\cite{wei2024cross}:} The Cross-temporal Contrastive Disentangling Model (CCDM) proposes a disentanglement-reconstruction strategy to separately enhance the spatial representations of classical and modern Chinese texts.
    \item \textbf{CGCLM~\cite{xiang2024cross}:} The Cross-Guidance Cross-Lingual Model (CGCLM) utilizes LLM to generate large parallel corpora of classical and modern Chinese text pairs. Additionally, it designs a mutual masking pre-training strategy to achieve semantic alignment between the two language styles.
    \item \textbf{Poka~\cite{wei2024knowledge}:} The Poetry Knowledge-augmented Joint Model (Poka) uses semantic knowledge graphs and a knowledge-guided mask-transformer to bridge the semantic gaps between classical and modern Chinese, enhancing both thematic and emotional understanding.
\end{itemize}

\subsection{Experimental Results}

\paragraph{Main Results} The model performance on two datasets is shown in Table~\ref{table:main result}. Among all baselines, our framework achieves the best performance. On the FSPC dataset, our method outperforms the second-best by 2.51\% in accuracy. On the CCPD dataset, our method yields an improvement of 1.19\% in micro F1 and 1.63\% in macro F1. The second-best methods on both datasets (CGCLM and Poka) both incorporate modern Chinese corpora, indicating the effectiveness of modern Chinese interpretations. Our model still outperforms them, and the above-mentioned margins come from audio and visual features and the tri-modal fusion with textual features.

We observe that although the CCPD dataset has more categories and appears to present a more challenging task, the performance metrics for the CCPD dataset are still significantly higher than those for the FSPC dataset. This suggests that the FSPC task may inherently be more difficult, as its labels, such as `implicit negative' and `implicit positive,' seem less straightforward compared to the clearer labels in the CCPD dataset, like `homesick' and `frustrated'.

\begin{table}[t]
	\centering
    \small
	\begin{tabular}{cc|ccc}
		\toprule
        \multicolumn{2}{c|}{FSPC} & \multicolumn{3}{c}{CCPD} \\
        \midrule
		Method & Acc & Method & Micro F1 & Macro F1 \\
		\midrule
        ChatGLM & 59.48 & ChatGLM & 89.50 & 75.59\\
        CDBERT & 60.25 & GreaseLM & 84.44 & 69.39 \\
        RAC-BERT & 61.40 & HiAGM-TP & 88.27 & 72.88\\
        CCDM & 61.45 & LCM & 90.19 & 80.27\\
        CCMC & 61.86 & KPT & 90.72 & 82.98 \\
        CGCLM & 62.34 & Poka & 94.65 & 85.69 \\
        \midrule
        Ours &  \textbf{64.85} & Ours & \textbf{95.84} & \textbf{87.32} \\
		\bottomrule
	\end{tabular}
	\caption{Performance comparison on two datasets (\%).}
	\label{table:main result}
\end{table}

\begin{table}[t]
	\centering
    \small
	\begin{tabular}{ccc}
		\toprule
		  & Textual Model & + Multimodal Framework \\
		\midrule
        RoBERTa & 59.05 & \textbf{62.35} (+3.30) \\
        ERNIE & 59.55 &  \textbf{63.80} (+4.25) \\ 
        ChineseBERT & 59.00 & \textbf{62.30} (+3.30) \\
        SikuRoBERTa & 59.90 &  \textbf{64.10} (+4.20) \\
        GuwenBERT &  60.10 & \textbf{64.85} (+4.75) \\
		\bottomrule
	\end{tabular}
	\caption{Performance comparison with different single-textual modality models on the FSPC dataset (\%).}
	\label{table:backbone}
\end{table}

\paragraph{Interchange Textual Modality With Other Models} We test the performance of our framework in combination with various single-textual modality models. The results on the FSPC dataset are presented in Table~\ref{table:backbone}. By introducing multimodal features, our framework improves the performance of all baselines, resulting in an average improvement of 3.96\% in accuracy. GuwenBERT (used in the main experiment) and SikuRoBERTa, both pre-trained on classical Chinese corpora, achieve the top two rankings.

\paragraph{Generated Visual Features vs. Image Features} We compare the effect of extracting visual features from real images (Baidu Images) of imagery words~\cite{su2023misc} versus those generated by text-to-image models. We replace the visual feature generation module ($+trans\&audio\&vision$) with image features ($+trans\&audio\&image$). 
The model using image features outperforms the one without visual modality ($+trans\&audio$), demonstrating the effectiveness of visual features.
However, it underperforms our model that uses generated visual features, indicating that our method provides more comprehensive visual features for the poetry.

\paragraph{Broader Dialect Exploration} To validate the effectiveness of dialect features, besides Mandarin and Cantonese, we also experiment with other southern dialects that we can find, including Wu (Suzhou), Minnan, and Chaozhou. As shown in Table~\ref{table:dialect FSPC}, on FSPC, incorporating these dialects alongside Mandarin yields an average Acc improvement of 3.08\% over the model without any audio features and 0.18\% over that using only Mandarin features. On CCPD, the corresponding F1 improvements are 0.70\% and 0.24\%.

\begin{table}[t]
	\centering
	\begin{tabular}{lc}
		\toprule
		Method & Acc \\
        \midrule
        \multicolumn{2}{c}{Ablation Study} \\
        \midrule
        $\text{GuwenBERT}$ & 60.10 \\
		$\text{GuwenBERT}_{+ trans}$ & 61.55 \\
		$\text{GuwenBERT}_{+ trans\&audio}$ & 64.00 \\
		$\text{GuwenBERT}_{+ trans\&audio\&vision}$ & 64.45 \\
		$\text{GuwenBERT}_{+ trans\&audio\&vision\&dialect}$ & \textbf{64.85} \\
        \hline
        \midrule
        \multicolumn{2}{c}{Comparison with Image Feature} \\
        \midrule
        $\text{GuwenBERT}_{+ trans\&audio\&image}$ & 64.30 \\
		\bottomrule
	\end{tabular}
	\caption{
    Ablation study results (top) and comparison of generated visual features and image features (bottom) on the FSPC dataset.}
	\label{table:abalation}
\end{table}

\begin{table}[t]
	\centering
    \small
	\begin{tabular}{cccccc}
		\toprule
		Method & Mand & Cant & Wu & Minnan & Chaozhou \\
		\midrule
		Single & +2.90 & +3.10 & +2.35 & +2.85 & +2.70 \\
		  + Mandarin & / & +3.30 & +3.05 & +3.00 & +2.95 \\
		\bottomrule
	\end{tabular}
	\caption{Improvement over our model without audio (61.55\% in Acc) using 5 different dialects on the FSPC dataset.}
	\label{table:dialect FSPC}
\end{table}

\begin{figure}[t]
	\centering
	\includegraphics[width=0.48\textwidth]{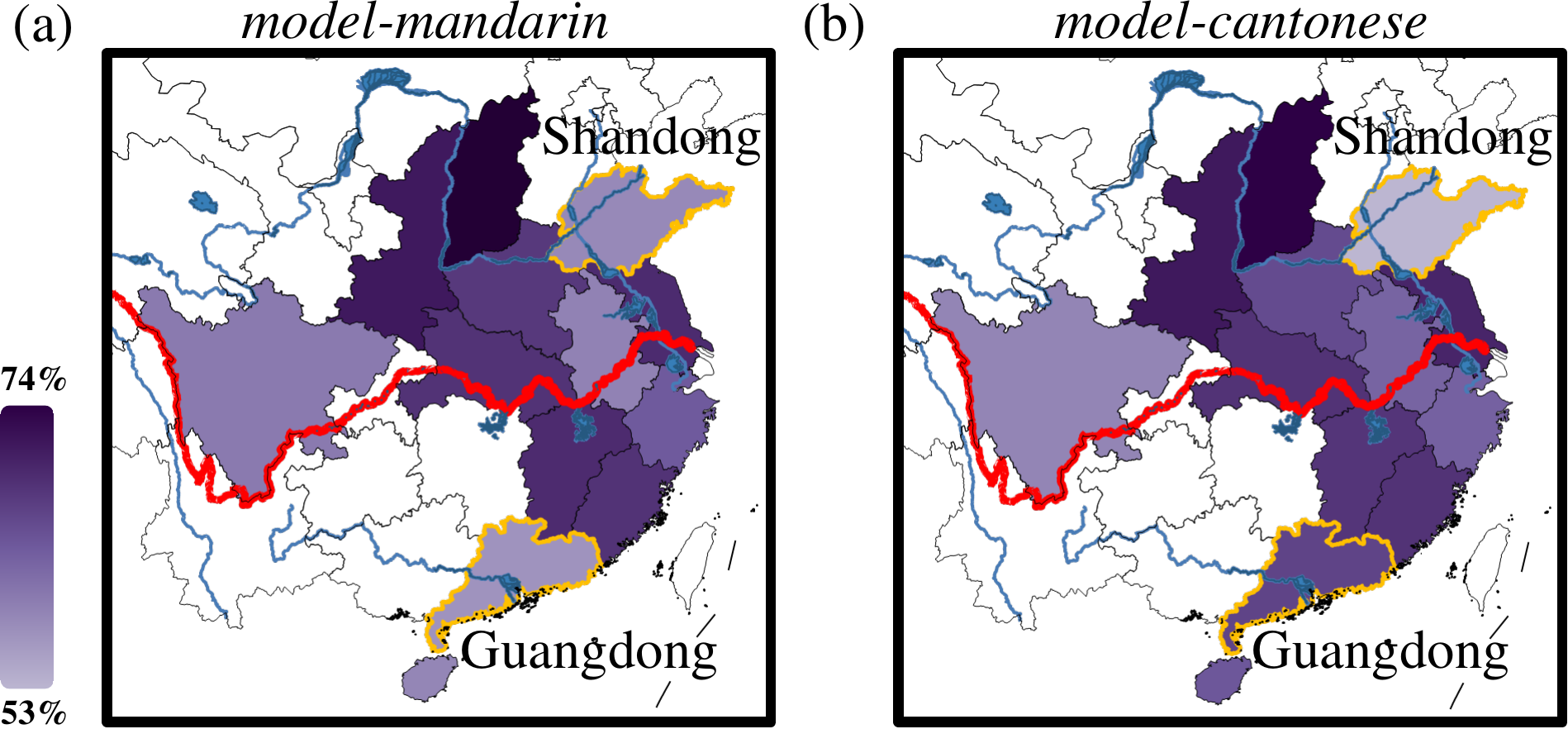}
	\caption{Comparison of \textit{model-mandarin} (in (a), trained on Mandarin audio) and \textit{model-cantonese} (in (b), trained on Cantonese audio) tested on poems from various provinces. The red line represents the Yangtze River, which divides northern and southern China.}
	\label{fig:discussion}
\end{figure}

\subsection{Ablation Study}

To validate the framework, we incrementally add our custom-designed modules to the GuwenBERT, including the LLM translation ($trans$), audio modality module ($audio$), visual modality module ($vision$) and dialect fusion module ($dialect$). Then we reconduct the experiment on the FSPC dataset. 

As shown in Table~\ref{table:abalation}, the results are in line with expectations. Among all modules, the audio features contribute the most significant improvement ($+trans$ vs. $+trans\&audio$, +2.45\%).
Following the audio modality module, the improvement brought by the LLM translation is notable ($\text{GuwenBERT}$ vs. $+trans$, +1.45\%), but still lags behind baselines specifically designed for aligning classical and modern Chinese texts ($+trans$ vs. CGCLM, -0.79\%). This suggests that the overall performance of our framework could be further enhanced by integrating more advanced single-textual modality methods. The visual modality module and the dialect fusion module contribute 0.45\% and 0.40\% in accuracy, respectively, which is less than the two modules above. A possible explanation is that visual features are more effective for certain poetic sentences that contain rich imagery words, and dialect features are relevant to poems from their corresponding regions.

\section{Southern Dialects for the Southern Poems?}

Does the use of southern dialects, such as Cantonese, significantly improve accuracy for southern poems, and northern dialects for northern poems? To investigate this, we analyze the regional effects of different dialect features on classification results. Specifically, we use Mandarin audio and Cantonese audio as inputs for the audio modality module, training two separate models: \textit{model-mandarin} and \textit{model-cantonese} on the FSPC dataset.

The two models exhibit similar overall accuracy (64.45\% for \textit{model-mandarin} vs. 64.50\% for \textit{model-cantonese}), but they differ in their performance on northern and southern poems. A poem is categorized as northern or southern based on the province of origin of its poet. For northern poems, \textit{model-mandarin} outperforms \textit{model-cantonese} by 0.94\%, while for southern poems, \textit{model-cantonese} outperforms \textit{model-mandarin} by 0.39\%. This trend is even more pronounced in certain representative provinces: for Shandong Province in the north, \textit{model-mandarin} surpasses \textit{model-cantonese} by 4.44\% (Shandong in Figure~\ref{fig:discussion}(a) vs. Shandong in Figure~\ref{fig:discussion}(b)), while in Guangdong Province, where Cantonese is most prevalent, \textit{model-cantonese} outperforms \textit{model-mandarin} by 8.11\% (Guangdong in Figure~\ref{fig:discussion}(b) vs. Guangdong in Figure~\ref{fig:discussion}(a)). The observed differences between the North and South, as well as between provinces, underscore the role of dialects as effective features for understanding classical poetry within specific regional contexts.

\section{Conclusion}
We have proposed a multimodal Chinese representation framework for sentiment analysis of classical Chinese poetry. This framework extracts phonetic features, integrates regional dialect audio, generates visual features, and merges them with LLM-enhanced textual features using multimodal contrastive representation learning. It achieves state-of-the-art performance on two public datasets and seamlessly integrates with single-textual modality methods. Further analysis shows that dialectal features significantly aid in classifying poems from their corresponding regions. For future work, we intend to explore specialized approaches for reconstructing ancient pronunciations, extend the framework to support fusing more than two dialects, and delve deeper into
sentence-level audio integration. Beyond classical poetry, we will expand our approach to a broader range of tasks in general Chinese representation and understanding.

\appendix




\bibliographystyle{named}
\bibliography{ijcai25}

\end{document}